\documentclass{article}

\usepackage{arxiv}

\usepackage[utf8]{inputenc} %
\usepackage[T1]{fontenc}    %
\usepackage[colorlinks,pdftex]{hyperref}       %
\usepackage{url}            %
\usepackage{booktabs}       %
\usepackage{amsfonts}       %
\usepackage{nicefrac}       %
\usepackage{microtype}      %
\usepackage{lipsum}
\usepackage{xcolor}
\usepackage{cleveref}
\usepackage{wrapfig}
\usepackage{placeins}
\usepackage{siunitx}
\usepackage[
    type={CC},
    modifier={by-nc-sa},
    version={4.0},
]{doclicense}

\usepackage{graphicx}                                       %
\usepackage{tikz}
\usepackage[font=footnotesize]{subcaption} %
\newcounter{daggerfootnote}

\crefformat{footnote}{#2\footnotemark[#1]#3}

\title{Occ-Traj120: Occupancy Maps with Associated Trajectories}

\author{
  Tin Lai\thanks{Both authors contributed equally} \\
  School of Computer Science\\
  University of Sydney\\
  Sydney, Australia \\
  \texttt{tin.lai@sydney.edu.au} \\
    \And
  Weiming Zhi\footnotemark[1] \\
  School of Computer Science\\
  University of Sydney\\
  Sydney, Australia \\
  \texttt{weiming.zhi@sydney.edu.au} \\
    \And
  Fabio Ramos\\
  School of Computer Science\\
  University of Sydney\\
  and NVIDIA USA\\
  \texttt{fabio.ramos@sydney.edu.au} \\
}

\begin{document}
\maketitle

\begin{abstract}
Trajectory modelling had been the principal research area for understanding and anticipating human behaviour.
Predicting the dynamic path by observing the agent and its surrounding environment are essential for applications such as autonomous driving and indoor navigation suggestions.
However, despite the numerous researches that had been presented, most available dataset does not contains any information on environmental factors---such as the occupancy representation of the map---which arguably plays a significant role on how an agent chooses its trajectory.

We present a trajectory dataset with the corresponding occupancy representations of different local-maps.
The dataset contains 400 locally-structured maps with occupancy representation and roughly around 120K trajectories in total.
Each map has few hundred corresponding simulated trajectories that navigate from a spatial location of a room to another point.
The dataset is freely available online\footnote{\label{dataset-url}Dataset available at \url{https://github.com/soraxas/Occ-Traj120}}.
\end{abstract}


\section{Introduction}

Trajectory datasets are the essential cornerstone in mobility data analytic and pedestrians motion predictions.
It is essential for the development of autonomous vehicles~\cite{lasota2017_SurvMeth} or extracting knowledge of human tendency to navigate within a particular environment~\cite{gonzalez2008_UndeIndi}.
Constructing a statistical model on the behaviour of trajectories within different forms of obstacles, e.g. pedestrians navigating within a building, can enable applications such as location-aware pedestrian trajectories prediction for autonomous driving~\cite{ridel2018_LiteRevi}, robots learning trajectories from human~\cite{kruse2013_HumaRobo} and route suggestions for indoor guidance system~\cite{chang2008_DesiImpl} for visually impaired people~\cite{kapic2003_IndoNavi}.
Having such a predictive model will enhance current existing methods for motion prediction.

To understand the behaviour of pedestrian under different environment, we need to have a comprehensive dataset that contains a wide range of sampled trajectories within different spatial structures.
Moreover, building structures or roads that we encounter in our daily life tends to have similar structural design framework---as practical structures will naturally look-alike.
The measurements of corridors or room dimensions, however, will be deviated based on each specific use case.
Therefore, it is intuitive that one can utilise learning techniques to perform reasoning and generalise it to the unseen environment, and to build a probabilistic model to predict the likely trajectories from pedestrians while they are walking along some structured road.
The currently available dataset in our research community, however, are very limited in terms of map diversity and their representations.
The most available dataset only contains trajectories that are captured in a fixed location, and without any occupancy representation of the map itself.
Therefore, such a limitation will restrict our ability to build a probabilistic model on how do pedestrian behave while encountering obstacles.

Our dataset is gathered for the purpose of enabling models to evaluate their predictive performance on trajectories that are generated within some structured maps.
The dataset contains multiple sets of locally-structured occupancy maps that have the same underlying architectural design, but with randomly deviated dimensions.
Within each map, there are few hundreds of simulated trajectories that navigates from a spatial point in the map to another, and our dataset contains around \num[group-separator={,}]{120000} trajectories and \num[group-separator={,}]{2700000} waypoints in total.
\doclicenseThis

\section{Related works}

This section briefly introduces currently available datasets in human dynamics and trajectory prediction community.
In contrast to common datasets in other research area, the representations of trajectories dataset are much more diverse.
For example, in traditional image processing from computer vision area, there is usually a comprehensive de facto dataset for evaluation (e.g. ImageNet~\cite{deng2009_ImagLarg}).
However, trajectories can represents a wide range of motions, and there exists a lot of datasets on different scenes as trajectories are highly conditional on the environment.
For example, trajectories can represent the motion of vehicles, humans, or any flying entities.
They can also represents something more abstract, for example, the travelling path of particles or the transitional dynamics within robots.
The diversity of datasets is due to the difficulty having a universal dataset for a wide range of scenarios.

\begin{figure}[b!]
    \centering
    \includegraphics[width=0.55\linewidth]{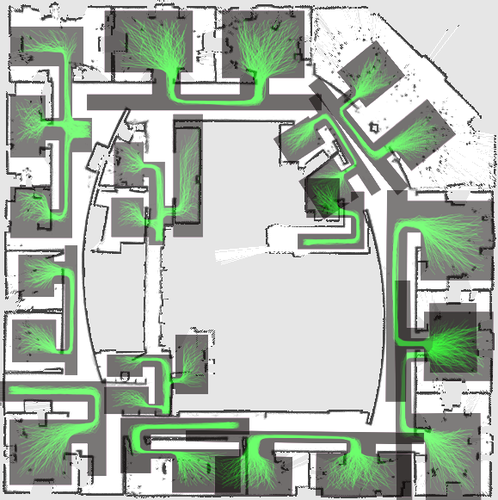}
    \caption{
        Multiple locally-structured maps overlaid with the Intel dataset.
        It demonstrates the decomposition of a floor-plan into multiple similar, but slightly deviated locally-structured maps.
        The semi-transparent black regions represent the free-space from the maps, and the green line-plot represent the map's corresponding trajectories in the dataset.
        \label{fig:intelmap_overlay}
    }
\end{figure}

Trajectory dataset in the form of videos are readily available, for examples, the ETH dataset~\cite{pellegrini2009_YoulNeve} and UCY Dataset~\cite{lerner2007_CrowExam} contain videos of pedestrians captured in a static street scenario, without other moving entities.
Videos from the dataset contain pedestrian tracking and works had been done on analysing crowd behaviour analysis.
The Stanford Dataset~\cite{robicquet2016_LearSoci} are videos captured from drones, and provides annotations for different kind of moving objects.
The VIRAT Video Dataset~\cite{oh2011_LargBenc} contains videos captured from
surveillance cameras, with the locations of cars, pedestrians, and other objects being annotated in the scene.
In additions, it also contains label of the activities in the videos.
However, none of the listed datasets contains the occupancy representations of the environment which limits its use case on learning to predict trajectories from maps.

The Edinburgh dataset~\cite{majecka2009_StatMode} contains pedestrian trajectories in the real-world, captured in the university of Edinburgh.
While the dataset is comprehensive and contain a wide range of human behaviour, it does not contains any occupancy information of the environment.
Hence, it limits the ability for statisaticalmodel to learn human behaviour based on surrounding environment.

\section{Trajectories Dataset}

In this section, we describe the generation of the locally-structured maps and the simulated trajectories within the maps.
We will then provide a statistical description of the proposing dataset and the distribution of each data point within the published data.

\subsection{Locally-structured maps}

\begin{figure}[tb]
    \centering
    \begin{subfigure}[t]{.5\linewidth}
        \includegraphics[width=\linewidth]{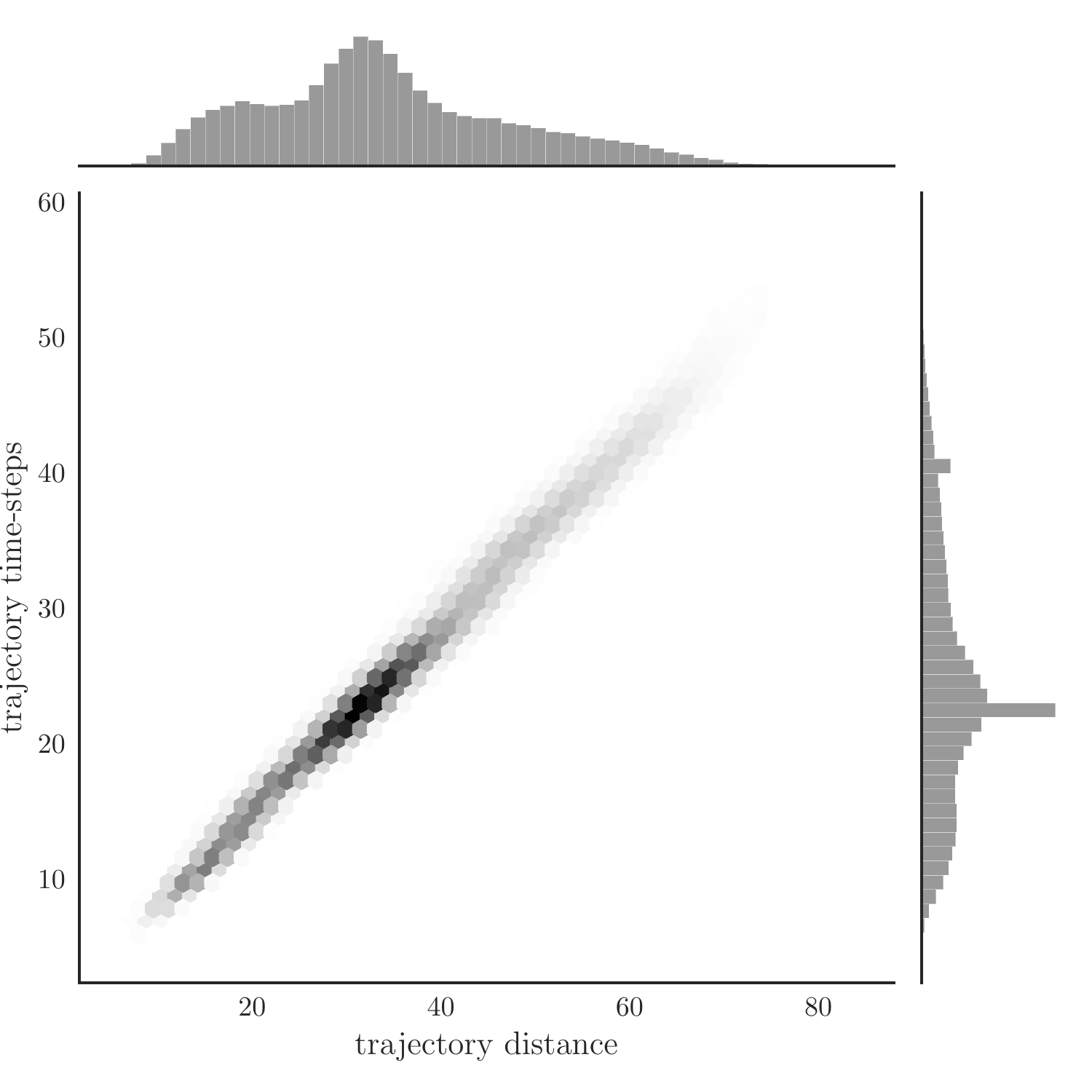}
        \caption{Paired distribution of trajectory length \label{fig:traj-length-timestep:paired-plot}}
    \end{subfigure}%
    \begin{subfigure}[t]{.45\linewidth}
        \includegraphics[width=\linewidth]{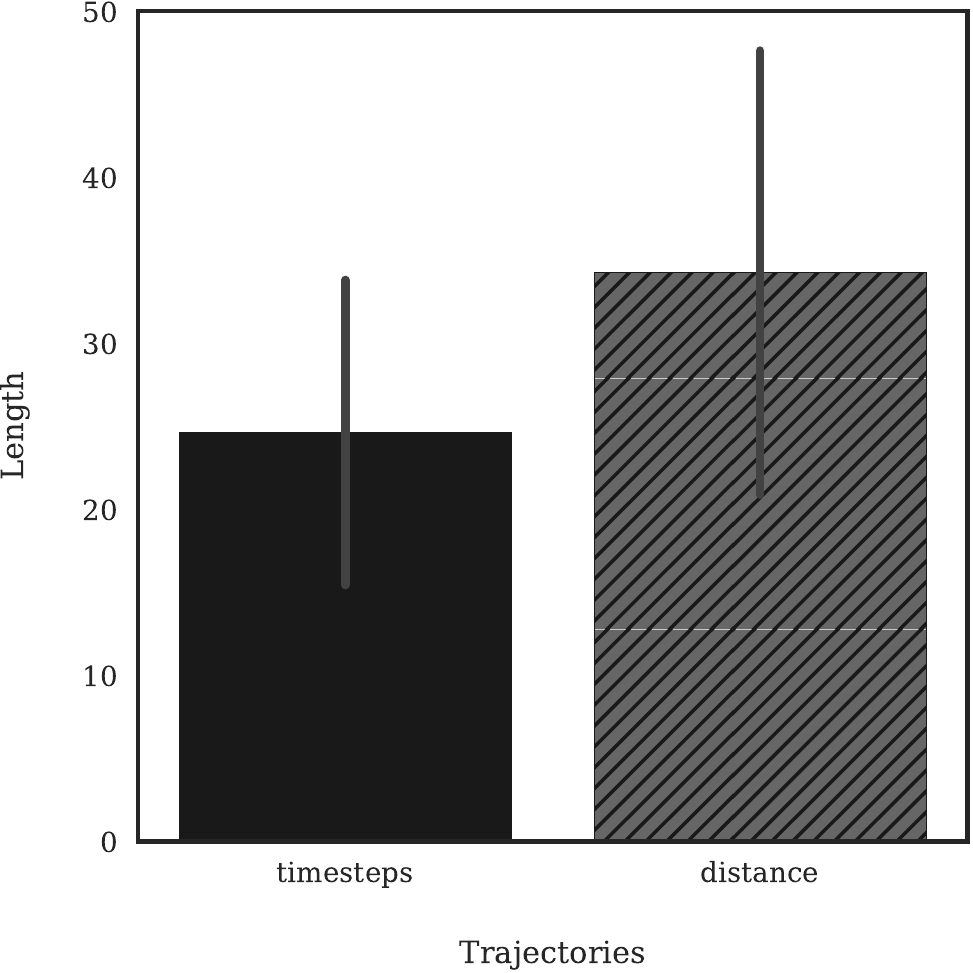}
        \caption{Overall trajectories' distance and time-steps \label{fig:traj-length-timestep:bar-plot}}
    \end{subfigure}
    \caption{
        Visualisation of the properties of the trajectories within the dataset.
        (\subref{fig:traj-length-timestep:paired-plot}) Shows the distribution of trajectory distance against time-steps.
        (\subref{fig:traj-length-timestep:bar-plot}) Summarised the overall length of the trajectories time-steps and distance.
        The error bar represents the standard deviation.
        \label{fig:traj-length-timestep}
    }
\end{figure}

This dataset is motivated by the desire to generating predictions of trajectories based on the occupancy representation of the surrounding environment.
A predictive model based on surrounding occupancy is useful as many buildings naturally tend to have common structures, and yet there is difficulty in obtaining a dataset that contains multiple representations of similar maps.
Moreover, it is immediate to realise that, for any given map, we can decompose it into smaller local-maps that are look-alikes.
That is, we can combine multiple smaller part of a map into an entire floor-plan, as illustrated by figure~\ref{fig:intelmap_overlay}.
The locally-structured maps can be re-orientated and scaled to be combined.

Therefore, in our dataset, it contains numerous locally-structured maps that are based on real floor-plans, e.g. the Intel dataset~\footnote{\texttt{http://ais.informatik.uni-freiburg.de/slamevaluation/datasets.php}}.
We used domain experts to categorise each sub-parts of a floor-plan into features that distinguish it from the others.
Then, each feature is allowed to be deviated from some mean value, while ensuring it falls within a certain range.
That is, the dimension of each distinguish feature will be sampled from $\mathcal{U}(a,b)$ for $0 < a,b < dim$ where $dim$ is the dimension of the local-map.
The choice of $a$ and $b$ will be dependent on the specific features and its overall layout of the local-map.
For example, the width of some corridor might be allowed to be deviated within some range, while ensuring it has enough space on its left for the corridor to be able to connect to the room structure of its left.

    \begin{wraptable}{r}{.43\textwidth}
     \caption{Description of the content and counts of the dataset}
      \centering
      \begin{tabular}{lll}
        \toprule
        \multicolumn{2}{c}{Stats}                   \\
        \cmidrule(r){1-2}
        Type       & Description            & Count       \\
        \midrule
        Map        & Total                  & 400   \\
                   & per type of local-map  & $\sim$100    \\
        Trajectory & Total                  & $\sim$120K  \\
                   & per each map           & $\sim$900   \\
        Waypoints  & Total                  & $\sim$2700K \\
                   & per each trajectory    & $\sim$40    \\
        \bottomrule
      \end{tabular}
      \label{tab:stats}
    \end{wraptable}

After the generation of different sets of locally-structured maps, we generate simulated trajectories with some motion planner.
Any motion planner that has the ability to perform planning in an occupancy map can be utilised.
In our settings, we used a modified RRT\textsuperscript{*}~\cite{karaman2011_AnytMoti} as our motion planner, as it can generate randomised trajectories in a reasonable time.
It is a popular sampling-based algorithm that had been researched extensively~\cite{elbanhawi2014_SampRobo}, embedded as a subroutine in a higher-level algorithm~\cite{lai2018_BalaGlob}, and there are extensions which incorporates learning-based approaches in this classical algorithm~\cite{ichter2018_LearSamp,bayeLocal}.
We modified the objective function such that instead of optimising the generated trajectory to have the shortest length, it will also account for the smoothness of the trajectory.
Start and end location of the trajectory are picked randomly from candidates, e.g. within the room structure or at the end of some corridors.
Precise coordinates of the start and end location are also randomised to simulate pedestrian starting and ending their path from different parts of the room.

\begin{figure}[tb]
    \centering
    \includegraphics[width=0.85\linewidth]{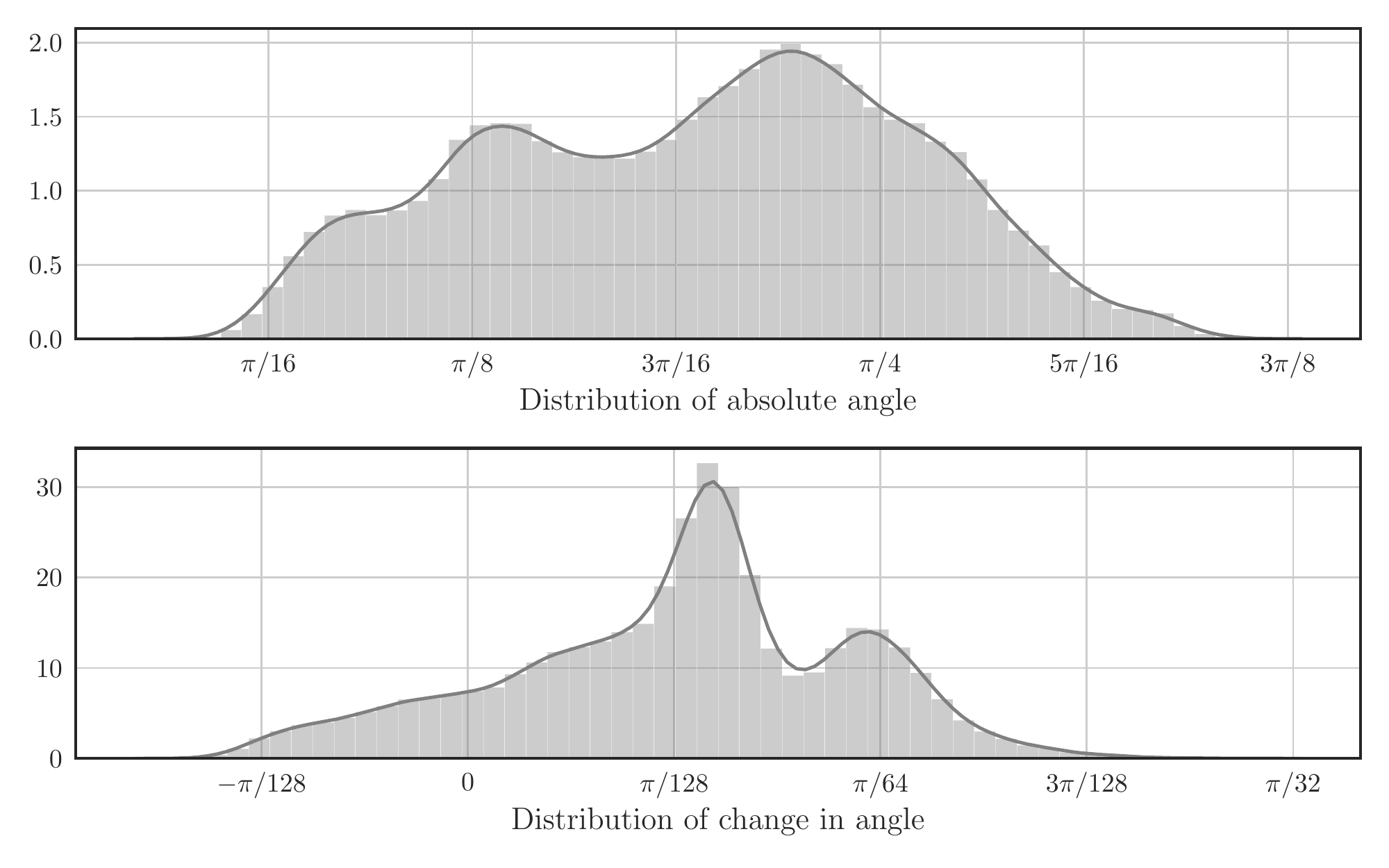}
    \caption{
        Distribution of the trajectories turning angle per time-step in the dataset
        \label{fig:angles-dist}
    }
\end{figure}

\subsection{Details on the dataset}

The dataset contains around \num[group-separator={,}]{120000} trajectories and 400 different maps in total, which is to the best of our knowledge the only trajectory dataset that contains a large amount of spatially different maps.
Table~\ref{tab:stats} summarise the amount of contents in our dataset.

\begin{figure}[bth]
\centering
\begin{tikzpicture}
\node (img) {
\begin{subfigure}[t]{.3\linewidth}
    \centering
    \includegraphics[width=\linewidth,clip]
    {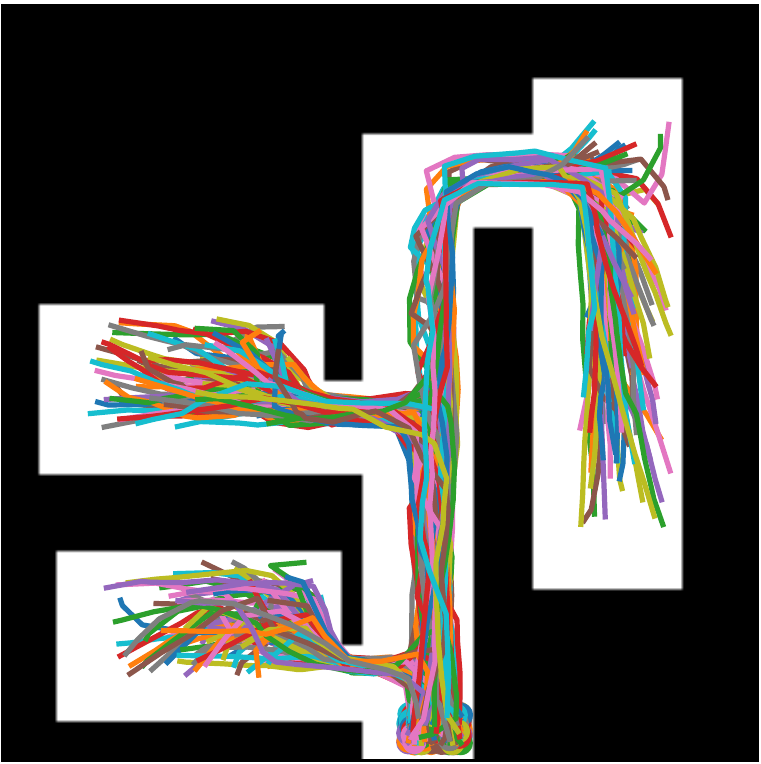}
    \includegraphics[width=\linewidth,clip]
    {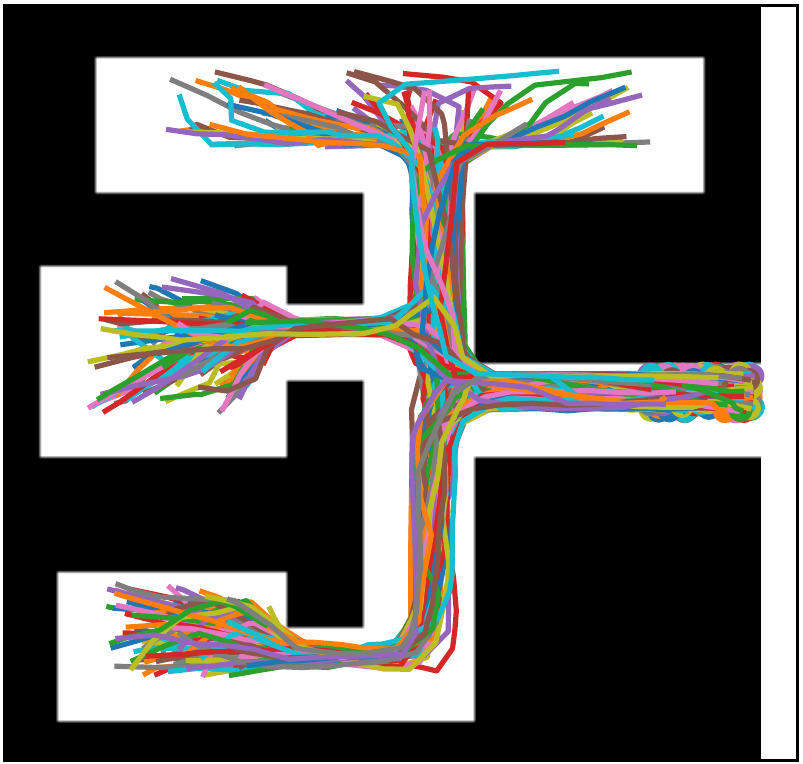}
    \includegraphics[width=\linewidth,clip]
    {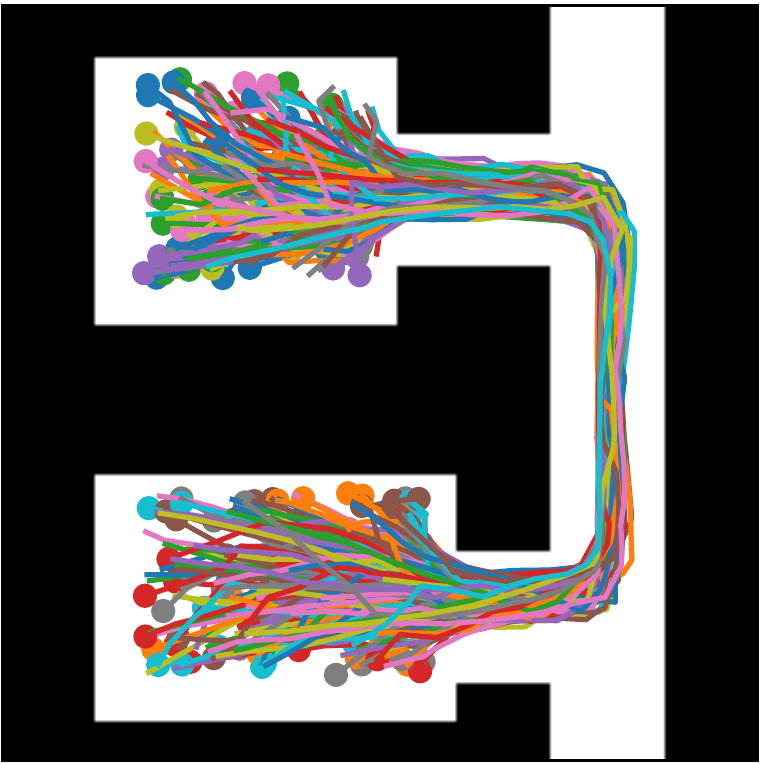}
    \includegraphics[width=\linewidth,clip]
    {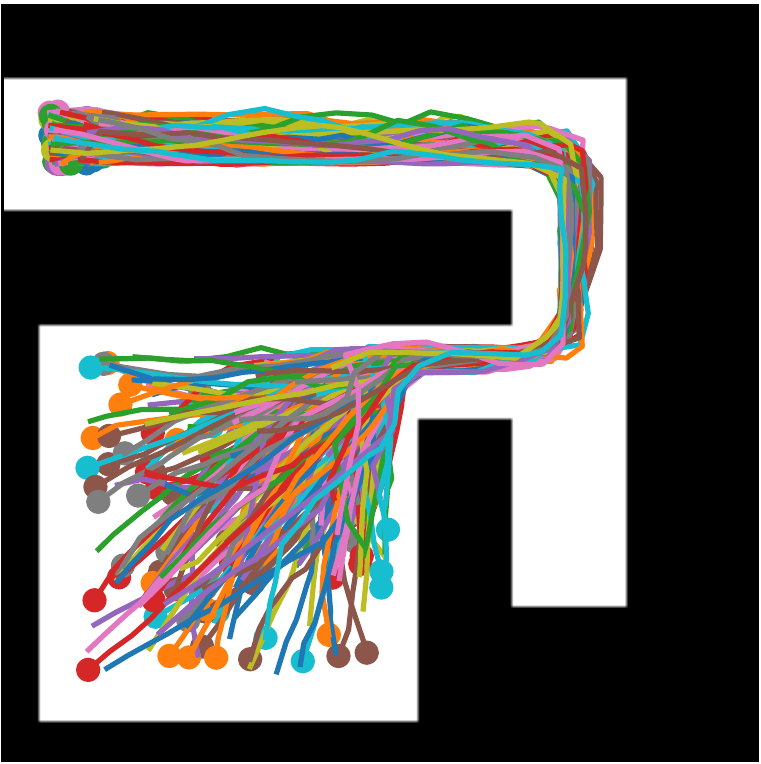}
\end{subfigure}%
\hspace{.4cm}
\begin{subfigure}[t]{.3\linewidth}
    \centering
    \includegraphics[width=\linewidth,clip]
    {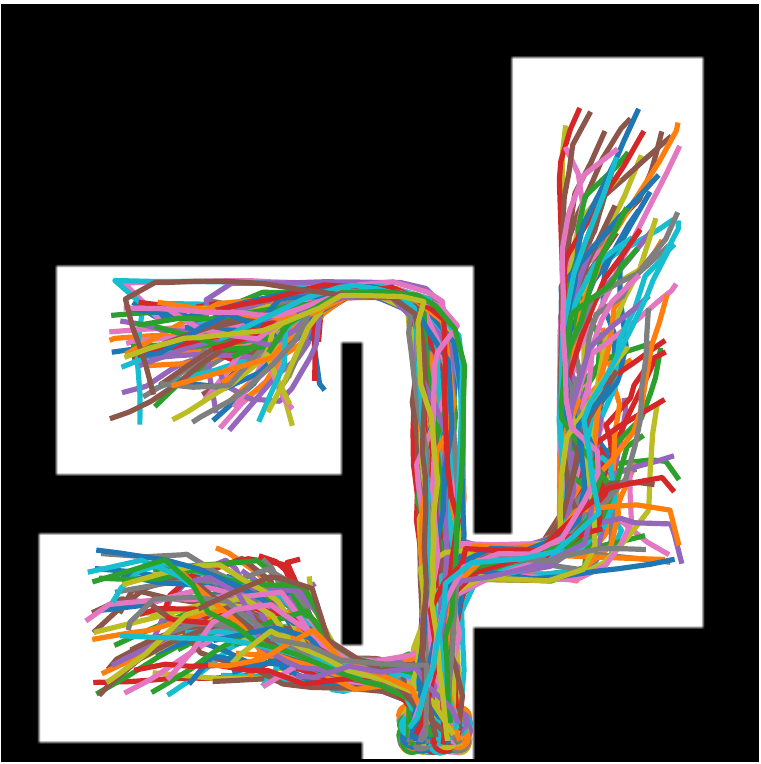}
    \includegraphics[width=\linewidth,clip]
    {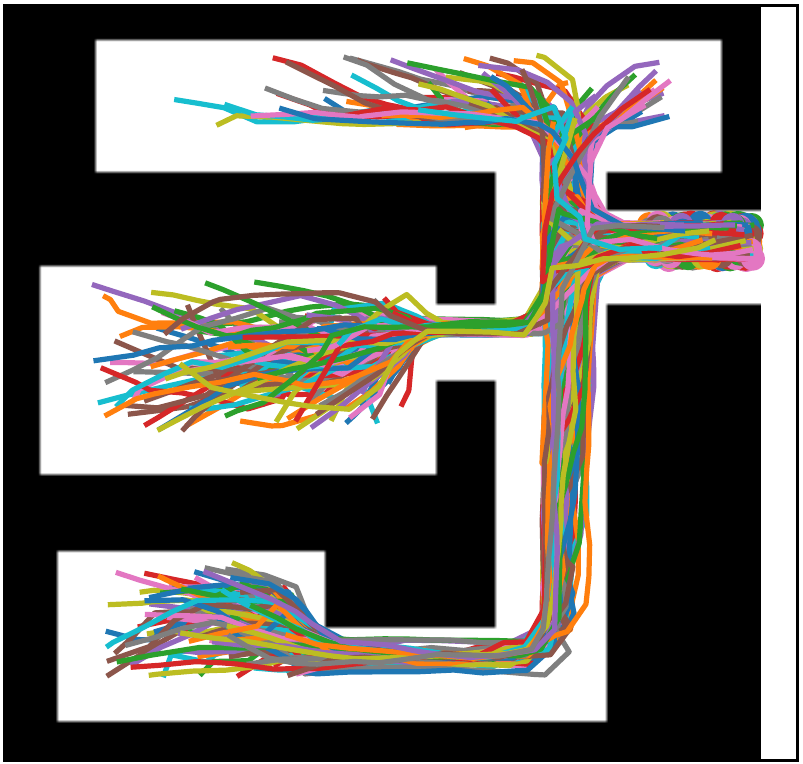}
    \includegraphics[width=\linewidth,clip]
    {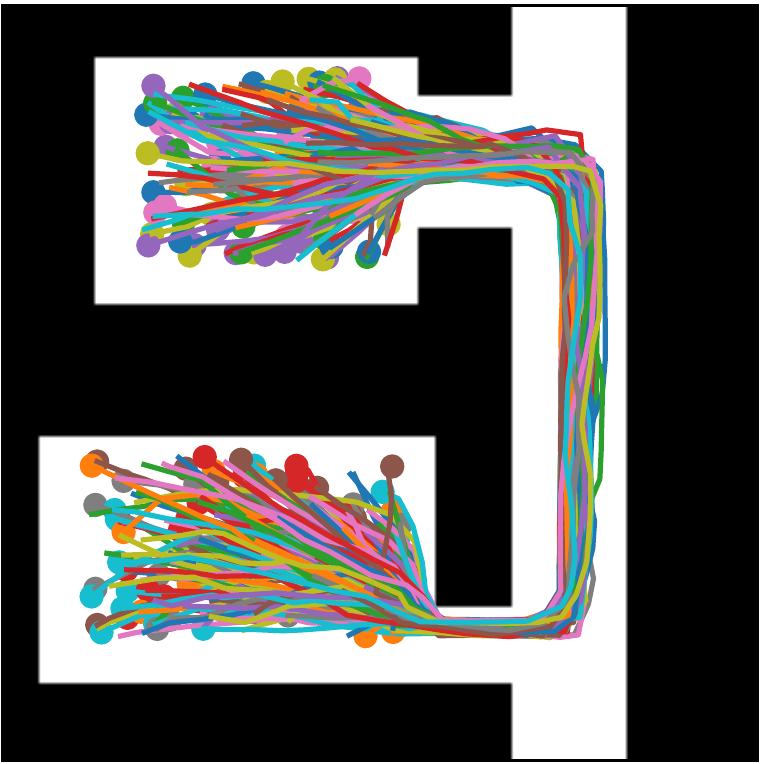}
    \includegraphics[width=\linewidth,clip]
    {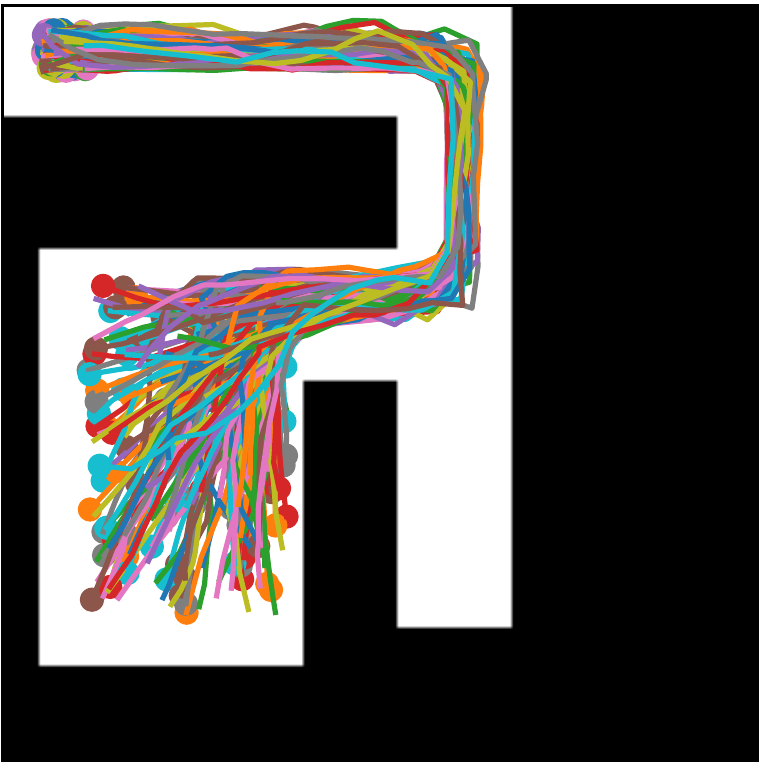}
\end{subfigure}%
\hspace{.4cm}
\begin{subfigure}[t]{.3\linewidth}
    \centering
    \includegraphics[width=\linewidth,clip]
    {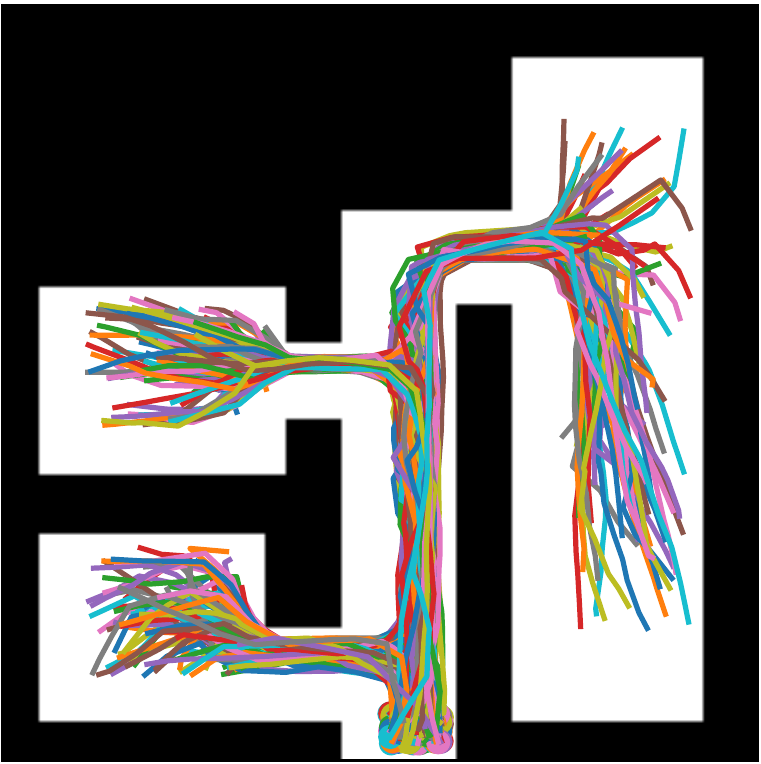}
    \includegraphics[width=\linewidth,clip]
    {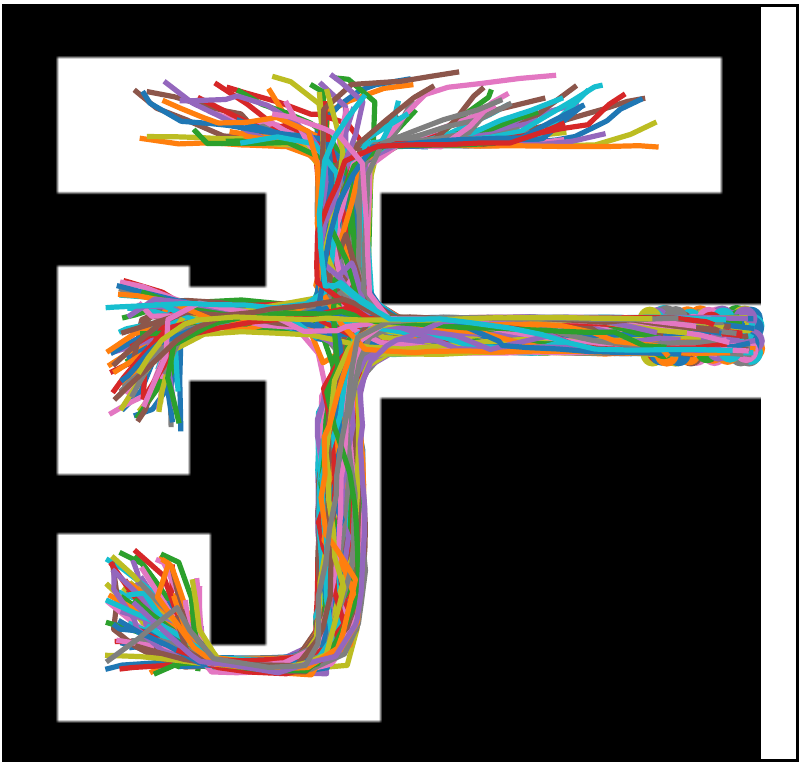}
    \includegraphics[width=\linewidth,clip]
    {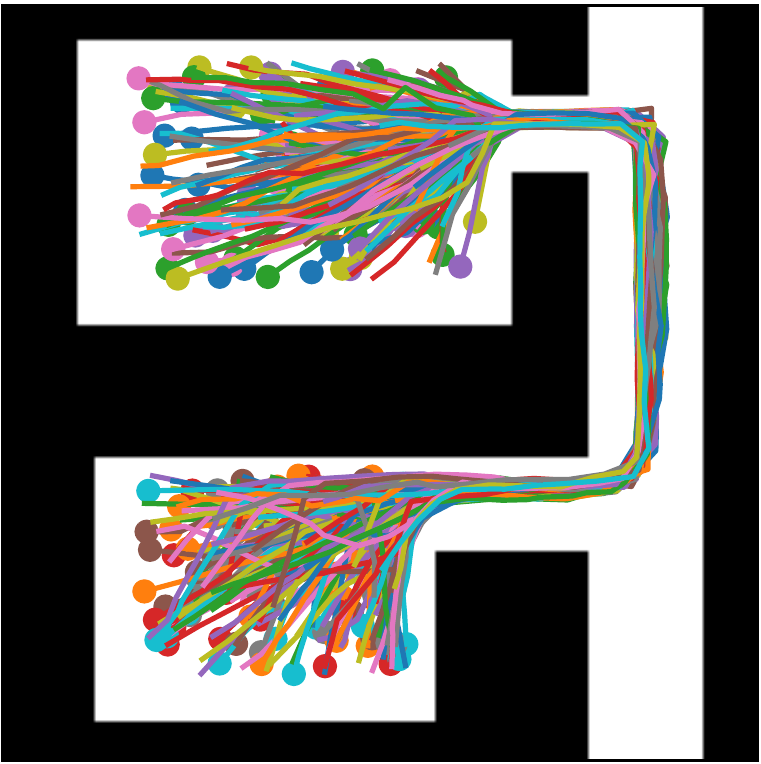}
    \includegraphics[width=\linewidth,clip]
    {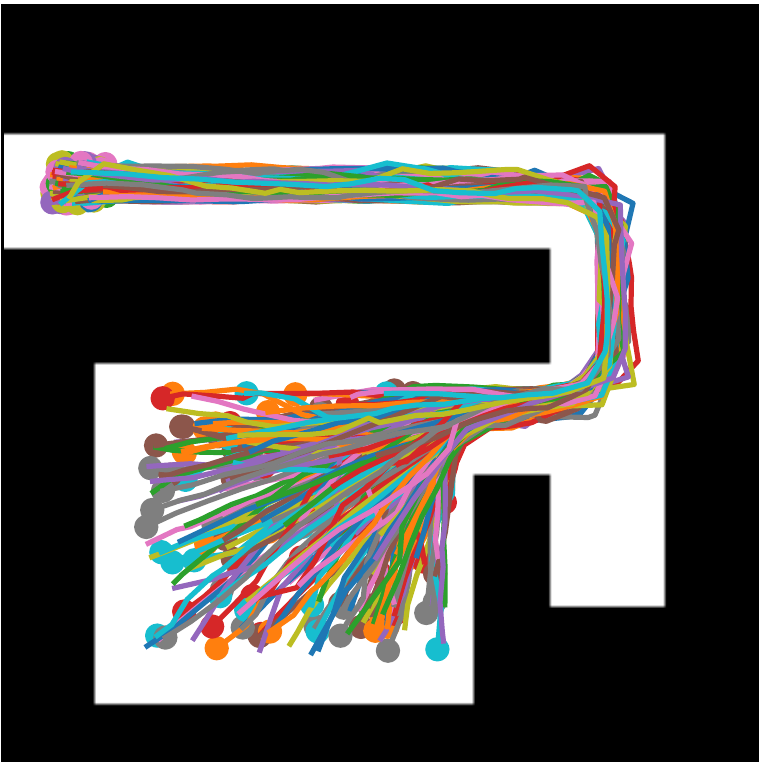}
\end{subfigure}%
};
\end{tikzpicture}%
    \caption{
        Examples of different type of maps and trajectories in the dataset.
        Each figures within the same row represents trajectories from similar local-maps, and each row demonstrates different kind of local-maps.
        Each trajectory are plotted with different colours, and the node at the end of trajectory represents the start point of that trajectory.
        \label{fig:examples-trajectories-plot}
    }
\end{figure}

\textbf{Distribution of trajectories length:}
There is a wide range of trajectories within a given map, and each one of them has a different number of time-steps and travel distance.
Figure~\ref{fig:traj-length-timestep} display the distribution of the trajectory length property.
There is a wide range of trajectories within a given map, and each one of them has a different number of time-steps and travel distance.
As shown by the figure, the number of trajectory time-steps is positively correlated with the trajectory length.
However, the dataset also contains a trajectory that might not be an optimal path and diverged from one other.
This fact is shown from the diverse range of distribution in figure~\ref{fig:traj-length-timestep:paired-plot}, and can be visualised by the plots of trajectories in figure~\ref{fig:examples-trajectories-plot}.

\begin{figure}[tb]
    \centering
    \includegraphics[width=0.75\textheight]{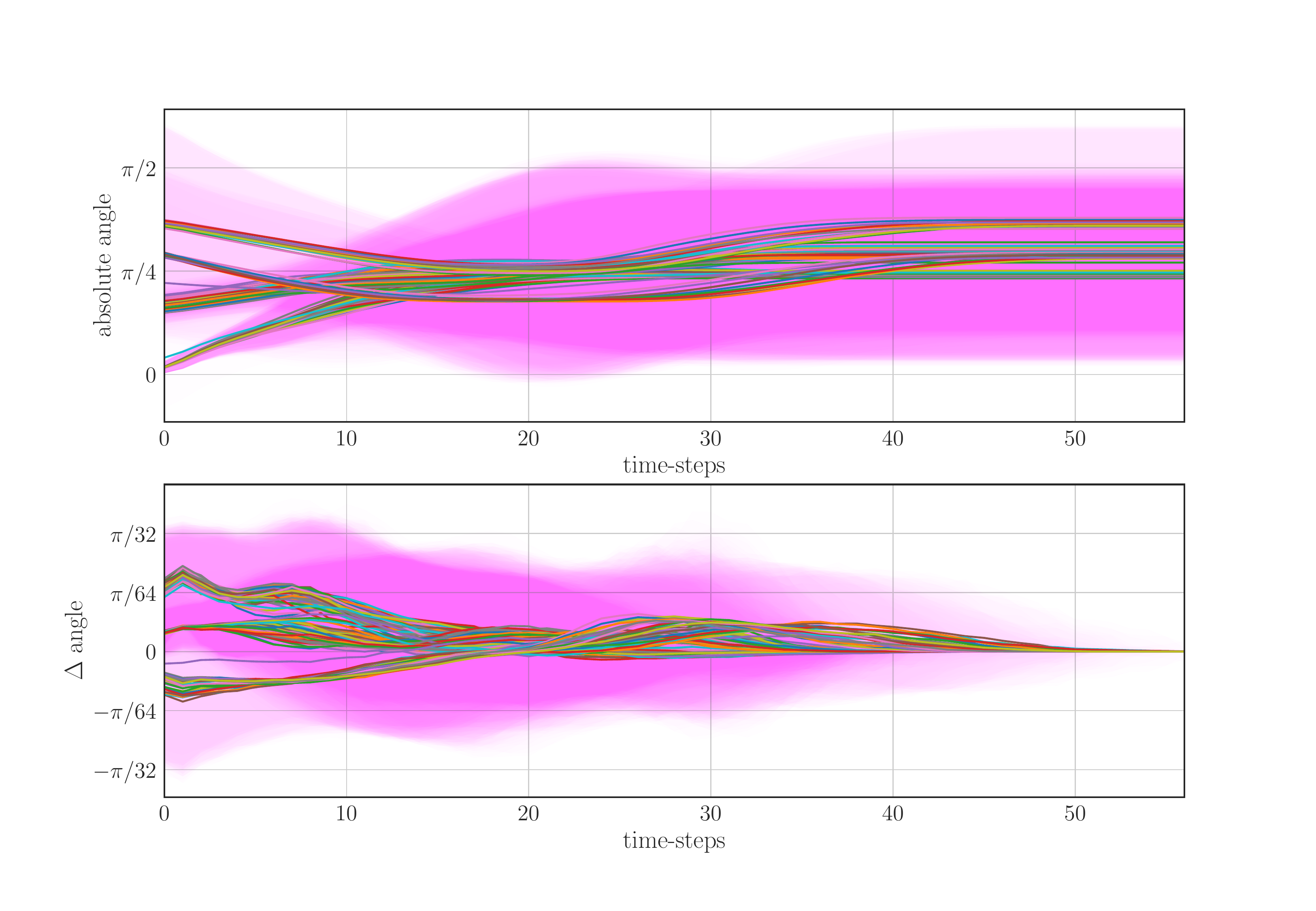}
    \caption{
        Plots of angles of trajectories against time-steps.
        The top figure shows the change of absolute angles of trajectory against each time-steps; the bottom figure shows the change in angle at each time-steps.
        For conciseness, the 120K trajectories are grouped together and the mean value is used for the plot.
        Shaded regions represents the standard deviations of the grouped trajectories.
        \label{fig:change-in-angles}
    }
\end{figure}

\textbf{Distribution of trajectories curvature:}
One of the main feature that can differ two similar trajectories are their curvature.
For any given pair of start and end location, the trajectories that connect those two locations can be varied in regards to the incremental curvature at each point.
Our dataset has a diverse set of trajectories that connects pairs of similar locations but with slightly deviated paths.
Figure~\ref{fig:angles-dist} displays the distribution of the angles and change in angle per each time-step.
There is a wide range of curvature along the trajectories, and the change in angle represents how sharp of a turn does each trajectory takes.
Figure~\ref{fig:change-in-angles} displays the angles of the trajectories plotted against time-steps.
The shaded regions denote the standard deviations of each grouped trajectories.
From the plots, it shows sets of trajectories navigate within the occupancy maps with different angles of curvature, and eventually reach to a stop after around $t>50$.

\section{Conclusion}

We presented \textsc{Occ-Traj120}---a dataset of occupancy maps with associated trajectories.
The proposed dataset addresses the issue of lack of data for machine learning technique to learn trajectory from associated occupancy maps.
The dataset contains representative trajectories and local-maps that are typical from common floor plans.
We had provide the dataset freely online\cref{dataset-url}, and presented the distribution and visualisation of the statistical information in this paper.
We belief the freely accessible data will our research community to build more powerful model to learn human behaviour based on surrounding environment.

\FloatBarrier

\bibliographystyle{unsrt}  
\bibliography{references}  %

\end{document}